\documentclass[10pt,twocolumn,letterpaper]{article}

\usepackage[pagenumbers]{wacv}              
\usepackage[accsupp]{axessibility}

\usepackage{graphicx}
\usepackage{amsmath}
\usepackage{amssymb}
\usepackage{booktabs}
\usepackage{enumitem}
\usepackage{arydshln} 

\newcommand{\OurMethod}{MixtureGrowth}

\newcommand{\OurMethodWithoutEns}{MixtureGrowth w/out fusion}

\newcommand*\samethanks[1][\value{footnote}]{\footnotemark[#1]}

\usepackage[pagebackref,breaklinks,colorlinks]{hyperref}

\usepackage[capitalize]{cleveref}
\crefname{section}{Sec.}{Secs.}
\Crefname{section}{Section}{Sections}
\Crefname{table}{Table}{Tables}
\crefname{table}{Tab.}{Tabs.}

\def\wacvPaperID{2323} 
\def\confName{WACV}
\def\confYear{2024}

\begin{document}

\title{MixtureGrowth: Growing Neural Networks by Recombining Learned Parameters}

\author{Chau Pham$^1$\thanks{Equal contribution}\quad Piotr Teterwak$^1$\samethanks[1]\quad Soren Nelson$^2$\samethanks[1]~~\thanks{Work done while at Boston University} \quad Bryan A. Plummer$^1$ \\
 Boston University$^1$ \qquad Physical Sciences Inc$^2$ \\
{\tt\small\{chaupham,piotrt,bplum\}@bu.edu}    \quad{\tt\small snelson@psicorp.com}
}

\maketitle


\begin{abstract}
  Most deep neural networks are trained under fixed network architectures and require retraining when the architecture changes. If expanding the network's size is needed, it is necessary to retrain from scratch, which is expensive.  To avoid this, one can grow from a small network by adding random weights over time to gradually achieve the target network size. However, this naive approach falls short in practice as it brings too much noise to the growing process. Prior work tackled this issue by leveraging the already learned weights and training data for generating new weights through conducting a computationally expensive analysis step. In this paper, we introduce \OurMethod{}, a new approach to growing networks that circumvents the initialization overhead in prior work. Before growing, each layer in our model is generated with a linear combination of parameter templates. Newly grown layer weights are generated by using a new linear combination of existing templates for a layer.  On one hand, these templates are already trained for the task, providing a strong initialization. On the other, the new coefficients provide flexibility for the added layer weights to learn something new. We show that our approach boosts top-1 accuracy over the state-of-the-art by 2-2.5\%  on CIFAR-100 and ImageNet datasets, while achieving comparable performance with fewer FLOPs to a larger network trained from scratch. Code is available at \url{https://github.com/chaudatascience/mixturegrowth}.

\end{abstract}

\section{Introduction}
\label{sec:intro}
\begin{figure*}[t]
    \centering
    \includegraphics[width=\linewidth]{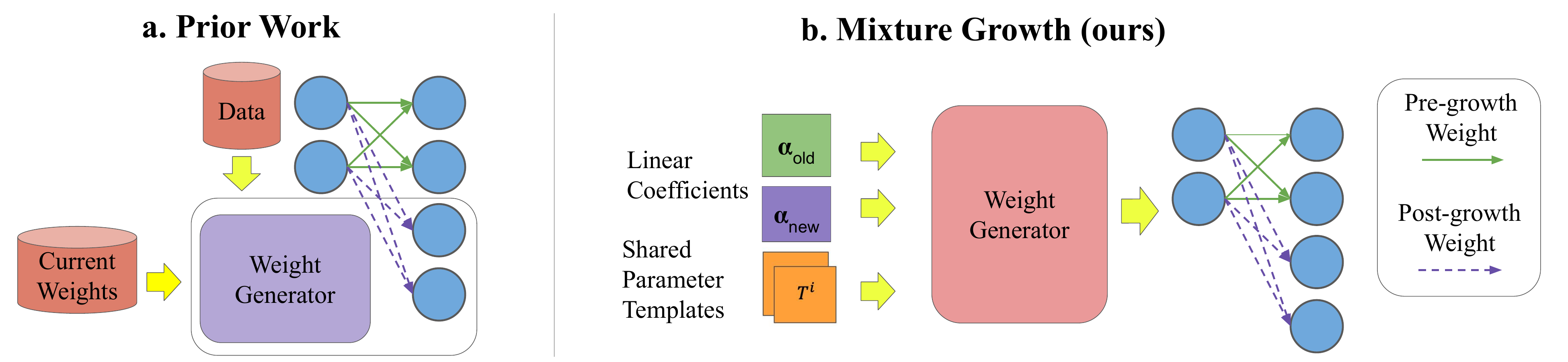}
    \caption{\textbf{A comparison of different strategies in growing neural networks.} \textbf{(a)} In prior work (\eg,~\cite{evci2022gradmax,wu2020firefly}), network growing methods  use loss-based local measures to initialize new weights at growth time. These methods depend on performing analysis using the training data and the current weights of existing layers, requiring significant computational overhead at the growth step; \textbf{(b)} In contrast, we investigate a template-mixing based method to grow the network. Given a small trained network, we start off with training another equally-sized network. Then, parameters of these small networks are shared across layers by template mixing schemes, where additional layer weights are generated using new linear coefficients for existing templates.}
    \label{fig:motivation}
 \vspace{-4mm}
\end{figure*}

Many tasks explore the trade-off between predictive performance and computational complexity, such as neural architecture search (NAS)~\cite{joglekar2020neural,bashivan2019teacher,tan2019mnasnet,dong2019one}, knowledge distillation~\cite{gou2020knowledge,hinton2015distilling}, and parameter pruning~\cite{hoefler2021sparsity,wang2021recent}. These methods result in a well-optimized model during inference but are often expensive to train (\eg,~\cite{tan2019mnasnet}). Thus, some researchers have explored an alternative procedure where one begins with a small network and increases its size over time (\eg,~\cite{evci2022gradmax,wu2020firefly}).  By first training a smaller and cheaper model, this approach can also be used to learn a target architecture with fewer floating point operations (FLOPs) than training the target network from the start.  The key challenge is deciding how to initialize any new layer weights after a growth step that avoids forgetting what the network has already learned while still leaving room for improvement. As illustrated in Fig.~\ref{fig:motivation}a, prior work explored methods that initialize new parameters through a computationally expensive analysis step that selects parameters that decrease the training loss~\cite{wu2020firefly} or increase gradient flow~\cite{evci2022gradmax}. 

Our paper seeks to find a solution for a scenario where we have a fully trained model and wish to expand it to a bigger one. 
To this end, we propose \OurMethod, a novel network growing framework that generates new weights by learning to reuse and combine parameters from the smaller network.  Our approach is inspired by recent success with template mixing methods~\cite{bagherinezhad2017lcnn,plummer2020shapeshifter,savarese2019learning,TeterwakSWE2022,YangCondCov2019}, where layer weights are generated using a linear combination of shared parameter templates.  Each layer obtains custom weights by using a unique set of linear coefficients, thereby increasing the expressiveness of the shared parameters.  The shared templates and the coefficients are trained jointly without altering the loss functions or network architecture for a target task.  When compared with traditional neural networks these methods have reported improved computational and parameter efficiency~\cite{bagherinezhad2017lcnn,plummer2020shapeshifter,TeterwakSWE2022}, reduced training time~\cite{bagherinezhad2017lcnn,plummer2020shapeshifter}, and improved task performance~\cite{plummer2020shapeshifter,savarese2019learning,TeterwakSWE2022,YangCondCov2019}.

\begin{figure*}[t]
\centering
\includegraphics[width=0.9\textwidth]{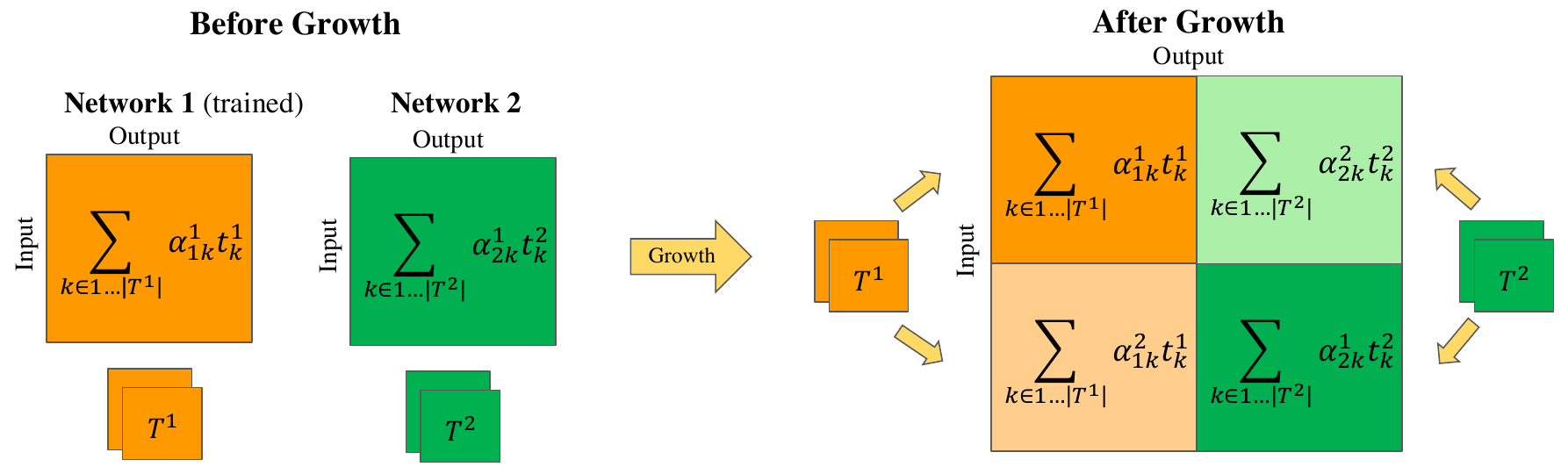}
\caption{\textbf{Illustration of the growth process in \OurMethod{} models where $g=2$ (\ie, double the size):} We explore how to grow models that have weights constructed from linear combinations of templates, which we call Template Mixing Models (see Sec.~\ref{sec:mixing}). Before growth, assume that one already has a fully-trained model (denoted as \textit{network 1}, left) and wishes to expand it. To this end, we first train another equally-sized model using a different set of templates (\textit{network 2}) for some number of epochs. Then, we merge the weights of the two networks so that their weights are tiled in the corners along the diagonal. In this example, we form a model 4x the size and resume training on the fully grown model (\textbf{target model}, rightmost). Quadrants that share templates but have different linear combination coefficients are presented as different shades of the same color. The squares correspond to layer weights whose input and output correspond to the input and output dimensionality. Note that all templates and alpha coefficients are learnable parameters.}

\label{fig:method}
\end{figure*}

A high-level overview of our approach is provided in Fig.~\ref{fig:motivation}b. 
Specifically, we postulate that since template mixing effectively shares parameters across layers, it can also be used to initialize new layer weights after a growth step. We investigate three research questions that were not explored in prior work.  First, template mixing provides a new mechanism for generating new layer weights after growing. For example, let us consider increasing the width of a layer $g$ times, resulting in most layers in the expanded network having $g^2$ times the weights of the original network.  Since the weights in the small network are generated using a linear combination of shared templates, we can generate the weights of the large network by initializing $g^2-1$ new sets of linear coefficients and concatenating the generated weights for the large layer (illustrated in Fig.~\ref{fig:method}). We find that careful initialization of these new linear coefficients is still necessary for the best performance, with some strategies, such as replicating the existing coefficients, having a detrimental impact on training.

The second research question we investigate is whether we should grow from a single small model, or if fusing two small models is better. We argue that training an extra small model provides more diversity in terms of learned features for the growth process. In a setting where we grow from a model that is half the width of the large one, then a small network would only take around 25\% of the FLOPs of the large network
making it relatively inexpensive.

The final research question we explore, following up on the previous question, is to find the optimal time to stop training the second network and start training a fully grown one. Given a fixed FLOPs budget, growing early gives more time to optimize all the layer parameters in the large network.  However, this also limits the ability of the second network, thereby raising the question of when to grow. We find that our approach is relatively insensitive to the exact training stage at which a growth step occurs, and could be used to grow late with minimal impact on task performance.

In summary, our contributions are: 
\begin{itemize}[nosep,leftmargin=*]
  \item We develop \OurMethod{}, a novel approach for growing a neural network over time by concatenating together weights of the old (small) network to fresh weights generated from newly initialized linear coefficients used to combine existing shared parameter templates.
  \item Our experiments show that \OurMethod{} is an effective growth method by outperforming the top-1 accuracy of prior work by 2-2.5\% on CIFAR-100 and ImageNet, demonstrating the effectiveness of our approach. 
  \item We analyze our approach for growing neural networks along axes such as sensitivity to growth point, the characteristics of learned features, and the effect of fusing two small models as opposed to growing from a single model. 
\end{itemize}

\section{Related Work}
\label{sec:related_work}

\noindent\textbf{Growing Neural Networks} has long been an active area of research; In early work, Ash \etal~\cite{ash1989dynamic} added neurons in single-layer systems, which later was extended by Fahlman \etal~\cite{fahlman1989cascade} to the deep network regime. It became common to pre-train neural networks greedily stacking Restricted Boltzmann Machines \cite{bengio2006greedy,hinton2006fast} to construct Deep Belief Networks \cite{hinton2009deep}. This was simplified by Vincent \etal~\cite{vincent2010stacked} with Stacked Denoising Autoencoders. More recently, Wen \etal~\cite{wen2020autogrow} proposed depth growing policies to automatically grow networks until performance stops improving. Maile \etal~\cite{maile2022and} proposed initializing new weights to maximize orthogonality. However, these works aim to automate network size selection by progressive growth, as opposed to minimizing the cost of training for a given network size.

In more relevant work, Wu \etal~\cite{wu2019splitting} and Chen \etal~\cite{chen2015net2net} replicated weights of the network to progressively increase its width, while Net2Net~\cite{chen2015net2net} leveraged function-preserving transformations to transfer knowledge when growing. These approaches replicate weights such that the pre-growth function is preserved, and then perturb or add noise to break symmetry to grow.  Firefly~\cite{wu2020firefly} also learned an overgrown network and then pruned uninformative weights. GradMax~\cite{evci2022gradmax} and NeST~\cite{dai2019nest} initialized weights to maximize gradient norms. Although these methods are more efficient than retraining from scratch, many of them still require a significant computational cost to analyze how the network should initialize the new network.
\vspace{5pt}\newline
\noindent\textbf{Template Mixing} improves upon hard parameter sharing, which directly reuses parameters across layers~\cite{collobert2008unified,zhang2014facial}, by representing layer weights as a linear mixture of shared templates (\eg,~\cite{plummer2020shapeshifter,bagherinezhad2017lcnn,savarese2019learning,TeterwakSWE2022,YangCondCov2019}). Savarese \etal~\cite{savarese2019learning} found they could use template mixing to boost performance by sharing across multiple layers within a single network. Isomorphic networks~\cite{sandler2019non,zhmoginov2021compositional} can push cross-layer parameter sharing further by constraining the network to have layers of identical shape and size. Plummer \etal~\cite{plummer2020shapeshifter} removed the need to have many identical layers by upsampling/downsampling templates when shared between layers of different shapes. Interestingly,  these methods exhibit similarities to the concept of modular learning, where network layers are shared across multiple tasks, as explored in prior research~\cite{pfeiffer2023modular, meyerson2018beyond}. While modular learning relies on shared layer weights tailored to individual tasks, template mixing, in contrast, combines shared templates to dynamically generate layer weights. In this paper, we expand on the study of template mixing by asking new research questions not addressed in prior work.  Specifically, prior work in template mixing assumes the network being trained retains the same size over time.  However, in this work, we leverage the unique structure of template mixing to generate new weights when expanding the network during training.
\vspace{5pt}\newline
\noindent\textbf{Shrinking Neural Networks} alters the size of a model during training typically aimed to increase its computational efficiency at test time. There are two broadly used methods for this: distilling and pruning neural networks. In neural network distillation, the goal is to transfer knowledge from a larger network to a smaller network. The technique, introduced in Hinton \etal~\cite{hinton2015distilling}, replaced one-hot labels with the predictions of a larger network (\ie teacher network). In this way, knowledge from the teacher network is transferred to the smaller network called student network. Examples of subsequent work include methods that distill ensembles into a single model~\cite{shen2020mealv2} and highlight the importance of the student and teacher seeing identical augmentations over long training periods~\cite{beyer2022knowledge}.  In parameter pruning, the goal is to find and remove unimportant weights (\eg,~\cite{lecun1989optimalmozer1988skeletonization,thimm1995evaluating,strom1997sparse}), resulting in a more computationally efficient network at a cost of increases training time as it assumes one has trained the fully parameterized network until convergence.  Contrary to these tasks, our goal is to reduce the training time of the large network by growing a neural network over time.


\section{Growing via shared parameter initialization}
\label{sec:method}

Given a pre-trained small neural network architecture $F^S$ with layers $\ell^S_{1,...,N}$, our goal is to generate the weights of a larger neural network architecture $F^L$ with layers $\ell^L_{1,...,M}$.  Following~\cite{evci2022gradmax,wu2020firefly}, we evaluate settings where networks have the same number of layers (\ie, $M=N$), even though our approach could be used to grow in depth as well~\cite{plummer2020shapeshifter,savarese2019learning}.  Instead, we vary in the width of each layer, \ie, $|\ell^L_i| > |\ell^S_i|$, where the operator $|\cdot|$ returns the number of weights in a layer.  In our experiments, set the size of the $F^L$ to be twice that of $F^S$, although other settings are possible with minor changes to layer weight construction (\eg, as in~\cite{plummer2020shapeshifter,TeterwakSWE2022}).  The task then is to generate weights for $F^L$ that enable it to converge quickly using what has been learned in $F^S$. Methods are ranked on both the computational efficiency of obtaining the fully trained model $F^L$ as well as the quality of the final solution measured as task performance. 

Prior work for growing neural networks relies on performing a computationally expensive per-layer analysis of the gradients or training loss to initialize any new parameters in $F^L$ (\eg,~\cite{evci2022gradmax,wu2020firefly}). We avoid this issue by learning how to generate new weights using a weight generator based on shared parameters.  Note that we refer to \emph{weights} as the matrix of numbers used by a layer operation, such as the filters used by convolutional layers, and \emph{parameters} as the matrices being optimized using backpropagation.  In traditional neural networks, weights and parameters typically refer to the same set of matrices, but in this paper, we generate layer weights using trainable parameters. The parameters are linear combinations of templates which are the results of a process we refer to as \textit{\textbf{template mixing}}~\cite{plummer2020shapeshifter,savarese2019learning,TeterwakSWE2022} which we briefly review in Sec.~\ref{sec:mixing}.  Next, we discuss our strategy for learning more expressive parameter templates using model fusion (Sec.~\ref{sec:fusion}).  Finally, we introduce a process of initializing any new weights needed by $F^L$ by tiling weights generated using the templates initially learned from our small network $F^S$ (Sec.~\ref{sec:tiling}).

\subsection{Generating weights with template mixing}
\label{sec:mixing}

Template mixing networks (\eg,~\cite{bagherinezhad2017lcnn,plummer2020shapeshifter,savarese2019learning,TeterwakSWE2022,YangCondCov2019}) generate layer weights by combining parameter templates and have shown they provide superior parameter efficiency and performance than traditional neural networks.  We use the setting where each parameter template $T^{1,..,t}$ is the same size as its corresponding layer in the small neural network $F^S$ (\ie, $|T^k_i| = |\ell^S_i|$), although parameter templates of different sizes could be also used~\cite{TeterwakSWE2022}.  Thus,  the weights of the $i^{th}$ layer, donated as $\ell^S_i$, are generated via,
\begin{equation}
    \ell^S_i = \sum_{k \in 1, \dots, t} \alpha_i^{k} T_i^k,
    \label{eq:comb}
\end{equation}
\noindent where $\alpha_i^{k}$ is a trainable parameter that controls the contribution of template $T_i^k$ in generating weights for layer $\ell^S_i$.

We begin by pretraining our small network $F^S$ using template mixing. Following~\cite{savarese2019learning}, we share templates between any pair of layers of the same size (with a maximum grouping of two layers in a row that share templates) for the new weights. For example, if $T_i$ and $T_{i+1}$ are used to generate weights for identical layers $\ell^S_i$ and $\ell^S_{i+1}$, then they refer to the same set of shared parameters, and they would not share templates with layer $\ell_{i+2}^S$, even if was the same size.  The total number of trainable parameters over a standard neural network only increases by the number of coefficients $\alpha_i^{k}$ used to generate the weights (resulting in a hundred or so extra parameters in our experiments during training). Both the coefficients $\alpha_i^{k}$ and the templates $T^i$ are learned jointly via backpropagation from the task loss, with no additional loss terms over a standard neural network used during training. 
Note that this weight generation process does introduce some overhead during training, but this increase is negligible, accounting for just 0.03\% of the FLOPs in a forward pass for a batch size of 64 samples~\cite{plummer2020shapeshifter}.  

\subsection{Learning robust templates for growing}
\label{sec:fusion}

A primary difference between our \OurMethod{} and prior work in template mixing (\eg,~\cite{bagherinezhad2017lcnn,plummer2020shapeshifter,savarese2019learning,TeterwakSWE2022,YangCondCov2019}) is that the architecture in our first training stage is not the final network architecture.  
Instead, we use a standard neural network procedure for learning a small network and then grow to a larger network, similar to prior work on network growth (\eg,~\cite{evci2022gradmax,wu2020firefly}). Prior work in network growth would initialize new trainable parameters using a heuristic built from the small network. This could result in unfavorable initialization as the expressiveness of the large network is much more significant (due to having many more weights) than the small network.  The methods of prior work in network growth could be seen as focused on trying to recover (some) of that additional expressiveness missing in the small network. In contrast, our approach assumes that many of the templates we have learned in $F^S$ will generalize to the new weights needed after growing to the target network $F^L$. 

We explore two variants of our method: one we grow directly from a single trained model and reuse the templates for new weights after growth, and another where we learn an extra model using an independent set of templates before growth. We call this second approach \textit{\textbf{model fusion}}, illustrated in Fig.~\ref{fig:method}. Our method is motivated by the idea that independently trained models learn substantially independent features \cite{lakshminarayanan2017simple,dietterich2000ensemble}. Thus, fusing two models creates a more robust feature representation, which in turn boosts performance. 
From a pair of small trained networks $F^{S_1}, F^{S_2}$, we use them to initialize and train our target network $F^L$, which we discuss in the next section.


\subsection{Generating weights for the target network}
\label{sec:tiling}

After obtaining the pair of pre-trained small networks $F^{S_1}, F^{S_2}$ using the approach in Sec.~\ref{sec:fusion}, the goal is to use these networks to construct the weights of the large network $F^L$.  From $F^S$, we want to grow it $g$ times larger to obtain $F^L$. Thus, most layers in the large network require $g^2$ times the weights.  The only exception is the first and last layers, which are only $g$ times the size as the dimensions of the input and outputs have not changed. Fig.~\ref{fig:method} illustrates a tiling for the weights of each layer where $g = 2$. 

Since our small networks have been trained using template mixing, we can generate new weights by using a new set of coefficients ($\alpha$ in Eq.~\ref{eq:comb}).  Therefore, if we need to generate $g^2$ tiles of weights for a layer, then we would learn $g^2$ tile-specific coefficients, but the templates would be shared with other tiles generated from the same small subnetwork.  Of these $g^2$ tile-specific coefficients, we can generate $g^2-2$ new tiles by reusing our existing small subnetworks $F^{S_1}, F^{S_2}$, but we must initialize the new coefficients. We investigate multiple strategies for initializing these coefficients that are described below.

\smallskip

\noindent\textbf{Random coefficients} simply initializes new coefficients randomly from a standard normal distribution. By jointly optimizing all coefficients and shared templates, the neural network finds new low-loss regions after growth. However, it remains uncertain whether a random initialization would substantially increase diversity for network growth.  
\smallskip

\noindent\textbf{Coefficient copying} is motivated by the idea that the entire subspace of weights spanned by a linear combination of templates does not correspond to low loss. Finding new low-loss combinations without performing a thorough analysis is challenging, so instead we copy the coefficients of the subnetworks $F^{S_1}, F^{S_2}$ as the initial starting point for the new weights. These copied coefficients are not shared, so although they begin as the same values, they are changed independently while training $F^L$.  This can be seen as providing a starting initialization that increases the response of each subnetwork $g$ times in every layer after growing. 
\smallskip

\noindent\textbf{Orthogonal coefficients} uses orthogonal coefficients when sharing templates to encourage diversity in the generating weights, introduced by Savarese~\etal~\cite{savarese2019learning}. In other words, when each set of $\alpha$ coefficients can be considered a vector, all such $\alpha$ vectors that are associated with the same templates are initialized to be orthogonal following~\cite{saxe2013exact}.

\section{Experiments}
\label{sec:results}

\noindent\textbf{Experimental Setup. } We assume a trained network is available to grow. While our method starts with a small template mixing model, the baselines employ small networks of their own settings. For a fair comparison, all small networks have the same FLOPs (25\% of the target network).
\smallskip

\noindent\textbf{Metrics.} We rank methods based on their top-1 classification accuracy averaged over three runs on all datasets.  In addition, we report the number of floating point operations (FLOPs) required for training a model. This metric compares computational costs for different models regardless of their hardware. Thus, it is used widely to evaluate the efficiency of a network~\cite{tan2019efficientnet,dosovitskiy2020image,ruiz2021anytime}.  For our experiments, we normalize computational complexity across methods by specifying a model train in the same FLOPs to ensure a fair comparison. We use fvcore library\footnote{\url{https://github.com/facebookresearch/fvcore}} to compute FLOPs. 
\smallskip

\noindent\textbf{Baselines. } We employ the following baseline methods.
\smallskip

\begin{itemize}[nosep,leftmargin=*]
\item\textbf{Random} initializes any new weights randomly during growth. This refers to generating the weights directly, rather than randomly initializing the coefficients used in template mixing networks (described in Sec.~\ref{sec:tiling}).
\smallskip

\item\textbf{Net2Net~\cite{chen2015net2net}} is based on function preserving initialization to initialize the new weights in a way such that the prediction is unchanged. This ensures the newly grown model can have the same ability as the small one preventing the drop in performance at growth point. 
\smallskip

\item\textbf{Firefly~\cite{wu2020firefly}} grows a network using pruning.  Specifically, they overgrow the network, optimize it for a few steps, and then prune the network by using a Taylor Approximation of the loss function to estimate the importance score of new neurons.  Note that our approach avoids this overhead by learning to generate new weights by reusing our shared parameters, making the growing step significantly more computationally efficient.
\smallskip

\item\textbf{GradMax~\cite{evci2022gradmax}} proposes an approach that seeks to maximize the gradients of new neurons. This is motivated by the observation that initialization with large gradients results in quick progress on the learning problem. GradMax maximizes the gradients of the loss with respect to new parameters. It solves this approximately with SVD at growth time, making the growth step computationally costly when compared with our approach.  We use the author's code in our experiments\footnote{\url{https://github.com/google-research/growneuron}}, which we also use as our implementation of the Random and Firefly methods.
\end{itemize}

\subsection{Datasets and implementation details}

\noindent See the supplementary for details not described below.
\smallskip


\noindent\textbf{CIFAR-10 and CIFAR-100}~\cite{Krizhevsky09learningmultiple} contain 60K images of 10 and 100 categories, respectively. Each dataset is split into 50K images for training and 10K images for testing. We use the WideResnet architecture~\cite{zagoruyko2016wide} to grow from a WRN-28-5  to WRN-28-10. We train the model for 200 epochs on a single GPU using stochastic gradient descent.
Following prior work in network growing~\cite{evci2022gradmax}, Batch Normalization~\cite{IoffeBatchNorm2015} is only used before each block for a fair comparison. 


\noindent\textbf{ImageNet}~\cite{deng2009imagenet} contains 1K categories with 1.2M images for training, 50K for validation, and 100K for testing.  We train ResNet-50 models~\cite{he2016deep} for 90 epochs with a learning rate 0.1, decayed by 0.1 at 30, 60, and 80 epochs. We use stochastic gradient descent with a momentum 0.9, batch size 256 with cross entropy loss.  Before growth, each layer has half the channels of the final network.

\subsection{Results}
\label{sec:results_2}
\setlength{\tabcolsep}{2pt} 
\begin{table*}[t]

\centering


\begin{tabular}{rlccc:ccc} 
\toprule
 & Method & CIFAR-10~\cite{Krizhevsky09learningmultiple} & CIFAR-100~\cite{Krizhevsky09learningmultiple}  &  FLOPs Norm & ImageNet~\cite{deng2009imagenet} &  FLOPs Norm\\ 
\midrule
 \textbf{(a)} & Large (Target network) & 96.88\% & 81.39\% &  1.00X  &  76.49\% & 1.00X \\ 
 & Large Match & 96.62\% & 80.95\% &  0.45X & 73.85\% & 0.35X \\ 
 & Small & 96.53\% & 80.29\% &  0.25X & 72.48\% & 0.25X  \\
  & Small Match & 96.60\% & 80.30\% & 0.45X & 72.89\% & 0.35X\\

  \midrule

  \textbf{(b)} & Random & 95.61\% & 79.08\% & 0.45X  & 71.47\% & 0.35X\\
 & Net2Net\cite{chen2015net2net} &95.47\% & 79.20\% & 0.45X  & 72.29\% & 0.35X\\
 & Firefly~\cite{wu2020firefly} & 95.62\% & 78.90\% &  0.45X  & 71.30\% & 0.35X  \\
 & GradMax~\cite{evci2022gradmax} & 95.73\% & 79.05\% & 0.45X & 71.73\% & 0.35X\\  
  & \OurMethod{} w/o fusion & 96.05\% & 79.44\% &  0.45X & 73.02\% & 0.35X\\
 & \textbf{\OurMethod\ (ours)} & \textbf{96.66\%} & \textbf{81.50\%} &  0.45X & \textbf{74.51\%} & 0.35X\\

 \bottomrule
\end{tabular}
\caption{Network growing comparison using top-1 accuracy averaged over three runs. \textbf{(a)} Performance of training the target (post-growth) (\emph{Large}) network architecture from scratch (WRN-28-10 for both CIFAR datasets and ResNet-50 for ImageNet), as well as the performance of the small (pre-growth) (\emph{Small}) architecture (WRN-28-5 for both CIFAR datasets and a ResNet-50 with half the channels as the large network for ImageNet);  \textbf{(b)} Performance of network growing methods, where our approach outperforms the current SOTA.}
\label{table:main}

\end{table*}

\begin{figure*}[tb]
\centering
\begin{subfigure}[t]{\columnwidth}
\centering
    \includegraphics[width=.9\textwidth]{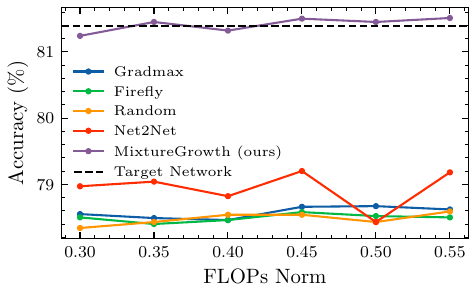}
      \caption{\textbf{Growing methods by FLOPs }}
  \label{fig:different_flops}
\end{subfigure}
\begin{subfigure}[t]{\columnwidth}
 \centering
  \includegraphics[width=.9\textwidth]{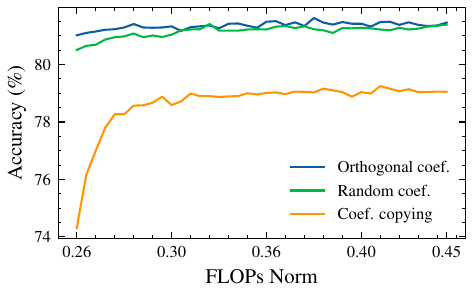}
  \caption{\textbf{Coefficient initializations}}
\label{fig:train_loss}
\end{subfigure}

\caption{Top-1 Accuracy on CIFAR-100. \textbf{(a)} Network growing methods under varying FLOPs. The dashed line represents the Target network's performance trained from scratch using a full training schedule.  We find \OurMethod{} not only outperforms Random, FireFly~\cite{wu2020firefly}, and GradMax~\cite{evci2022gradmax}, but also outperforms the target network with half the FLOPs;
\textbf{(b)} Performance of different coefficient initialization methods (Sec.~\ref{sec:tiling}) from growth point. See Sec.~\ref{sec:discuss} for discussion.}
\end{figure*}

Tab.~\ref{table:main} reports the performance of growing a neural network on CIFAR-10, CIFAR-100, and ImageNet.  
We see that \OurMethod{} is able to get better performance compared with training the target architecture from scratch (\textit{Target network}) with half of the FLOPs on CIFAR-100 dataset. As for ImageNet, our approach gets comparable performance with Target network using just one-third of the FLOPs. When considering the Target at the same FLOPs, it outperforms \textit{Large Match} (\ie the \textit{Target Network} but trained with the same FLOPs as network growing baselines) by a significant margin. Thus, our approach is capable of utilizing the existing small models to obtain a high-performing model in less time than a standard network.  

Prior work on network growth (Tab.~\ref{table:main}b) struggles to boost performance after growing when compared to simply training a regular small network (Tab.~\ref{table:main}a), denoted as \textit{Small} and \textit{Small Match} (\ie the \textit{Small} network but trained with the same FLOPs as network growing baselines). Note that compared to prior work, we explore larger network architectures in our experiments (\eg, a ResNet-50 on ImageNet in our work vs.\ a VGG11~\cite{simonyan2014very} evaluated in~\cite{evci2022gradmax}), suggesting these prior methods find generalizing to larger networks challenging. In contrast, our method can allow the model to gain improvement over growth and even surpass the target model on CIFAR-100 with half of the FLOPs. When compared with other growing methods, our \OurMethod{} approach gets around 1\% higher performance on CIFAR-10, 2.5\% on CIFAR-100, and 2\% on ImageNet. 

Fig.~\ref{fig:different_flops} demonstrates the efficiency and effectiveness of our method compared with four baselines: Random~\cite{berner2019dota}, Net2Net~\cite{chen2015net2net}, Firefly~\cite{wu2020firefly}, and Gradmax~\cite{evci2022gradmax} across various FLOPs budgets. Using WRN-28-10 as the target network, our method consistently outperforms the baselines with the same budget while being on par with or even outperforming the target network with half of the FLOPs. 

Tab.~\ref{table:gradmax_schedule} compares growing in a single step using \OurMethod{} to growing many times with the long growth schedule used by prior work~\cite{evci2022gradmax}. In this setting, not only do we obtain a significant boost in performance compared with the current state-of-the-art, but do so in half the FLOPs.  The additional FLOPs required by prior work are due to their growth overhead  (discussed at the beginning of Sec.~\ref{sec:results}).

\begin{table*}
\setlength{\tabcolsep}{6pt}
\centering
\begin{tabular}{lcccccccc}
\toprule &  \multicolumn{8}{c}{Second Network Training Time (FLOPs) } \\
\cmidrule { 2 - 9 } Dataset & 0.0 & 0.4 & 0.5  &  0.6 & 0.7 & 0.8 & 0.9 & 1.0 \\
\midrule CIFAR-100 &79.24\% & 80.16\%  & 80.46\% &81.08\% &80.9\% &81.09\% &\textbf{81.44}\% &81.27 \\
ImageNet & 73.02\% & 73.74\%  & 73.75\% &74.09\% &74.16\% &\textbf{74.51}\% &74.33\% &74.31\\
 \bottomrule

\end{tabular}

\caption{\textbf{Ablation on growth points} where we stop training the second network to train the full model. The first column is the extreme case where we grow without training the second network (\ie \OurMethodWithoutEns{}). The rest shows \OurMethod{}'s performance on varying growth points. \textit{Second Network Training Time} indicates the proportion of full-training FLOPs that was used to train the second network, \eg, 0.5 means we train the second network half of its full train before growth. We find that late growth where we almost fully train the second network before growing gains the best performance (boldface). All runs report top-1 accuracy at the same 0.35X FLOPs.} 


\label{table:when_to_grow}
\end{table*}

\subsection{Discussion}
\label{sec:discuss}

\noindent\textbf{How should we initialize linear coefficients?}  The first research question we explore is how to effectively initialize the coefficients used to generate new layer weights after growing.  In Sec.~\ref{sec:tiling} we describe three different methods for initializing these new coefficients.  We report their performance in Tab.~\ref{table:coef_init}, where we find orthogonal initialization, which helps promote the most diversity in the generated weights, outperformed both random initialization and coefficient copying. Notably, random initialization also performs well and approaches the performance of orthogonal initialization. Note, for coefficient copying's results, we also tested adding some small random noise to break the symmetry, but it did not substantially affect performance.
\smallskip

To obtain insights into how coefficient initialization methods perform at the growth step, we plot top-1 accuracy starting from the growth step until the end in Fig.~\ref{fig:train_loss}. We find that both orthogonal and random can maintain task performance despite being provided with new weights at the moment of growth and start improving performance with the large network. However, coefficient copying struggles at the growth step as its performance drops significantly. We suspect this may be due to the redundancies caused by reusing and replicating the same coefficients, making it difficult to learn new informative weights.
\smallskip
\setlength{\tabcolsep}{4pt} 
\begin{table*}[t]
\centering
\begin{minipage}{.48\linewidth}

\begin{tabular}{rlcc} 
\toprule
 & Method & Top-1 Acc.  &  FLOPs Norm   \\
 \midrule
 \textbf{(a)} & Random & 79.12\% &  0.77X \\
 & Firefly~\cite{wu2020firefly} & 78.94\% &  0.77X \\
 & GradMax~\cite{evci2022gradmax} & 79.24\% &  0.77X \\  
 \midrule
  \textbf{(b)} & \textbf{\OurMethod{}} & \textbf{81.50\%}   & 0.45X \\
 \bottomrule
\end{tabular}

\caption{\textbf{Network growing comparison on CIFAR-100}. \textbf{(a)} Using the long growing schedule used by the authors of GradMax; 
\textbf{(b)} \OurMethod{} with single growth for reference.}

\label{table:gradmax_schedule}
\end{minipage}
\hfill
\begin{minipage}{.48\linewidth}
\setlength{\tabcolsep}{8pt} 
\centering

\begin{tabular}{lc} 
 \toprule
 Method & Top-1 Accuracy    \\ [0.5ex] 
 \midrule
 Random coefficients  & 81.41\%  \\ 
 Coefficient copying & 79.05\% \\ 
 Orthogonal coefficients & \textbf{81.50}\%   \\
  \bottomrule
\end{tabular}

\caption{\textbf{Ablation on coefficient initialization methods} (described in Sec.~\ref{sec:tiling}) for \OurMethod{} on CIFAR-100. Orthogonal initializations perform best, suggesting the importance of having independent instantiated weights.
}

\label{table:coef_init}

\end{minipage}


\end{table*}

\noindent\textbf{Is growing by fusing two small models more effective than growing from a single model?}  The second question we explored was whether we could take advantage of having two small networks during growth.  Having a pair of networks would improve robustness as discussed in Sec.~\ref{sec:fusion}. Tab.~\ref{table:main} compares growing with and without model fusion (last 2 rows), where we find that fusing two models gains a consistent boost after normalizing by FLOPs.
\smallskip

\noindent\textbf{What is the right point during training to grow?} The previous question sheds some light on the importance of having a second trained model. However, how long we should train the second network is remain open. If we train the second one for a long schedule, then there is not much budget left to optimize the parameters of the large network. Conversely, if we grow it too early, then the second network is not sufficiently trained, which in turn may lead to a drop in overall performance. From this perspective, we can consider \OurMethodWithoutEns{} as an extreme case of early growth where we do not train the second network but instead grow to a full network directly. Tab.~\ref{table:when_to_grow} shows the performance of different growth points under the same FLOPs budget on CIFAR-100 and ImageNet datasets. We observe that late growth where the second network was almost fully trained gives the best performance. This confirms our assumption that growing with two networks is more robust.


\subsection{Feature analysis}
\label{subsubsec:feature_analysis}

\begin{figure}[t]
\centering
\begin{subfigure}[t]{0.48\columnwidth}
    \centering
    \includegraphics[width=\textwidth]{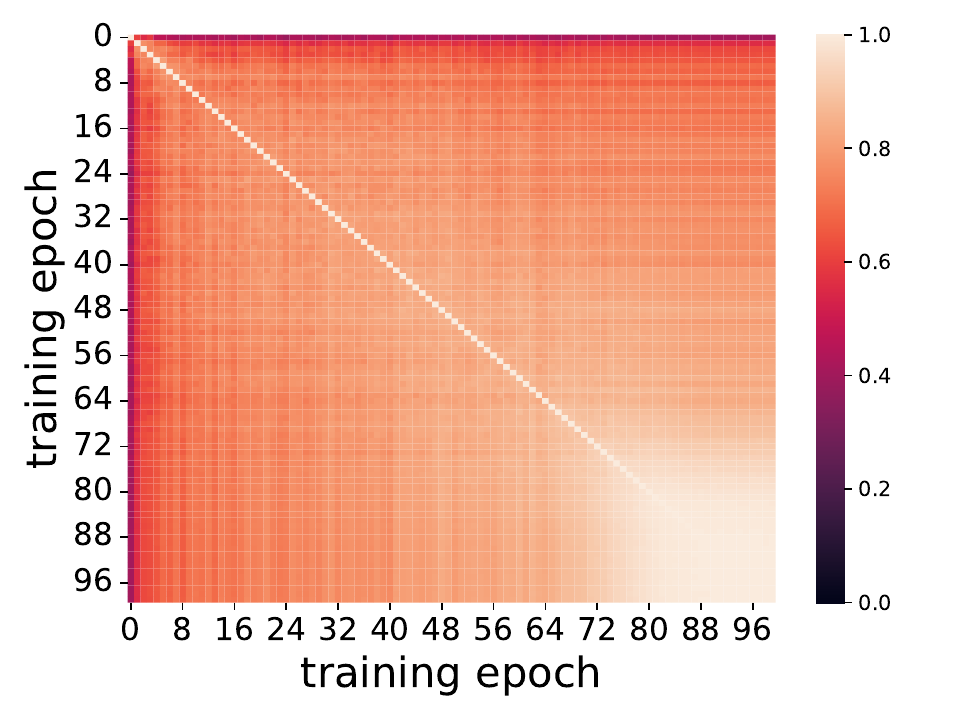}
    \caption{}
    \label{fig:t}
\end{subfigure}
\begin{subfigure}[t]{0.48\columnwidth}
    \centering
\includegraphics[width=\textwidth]{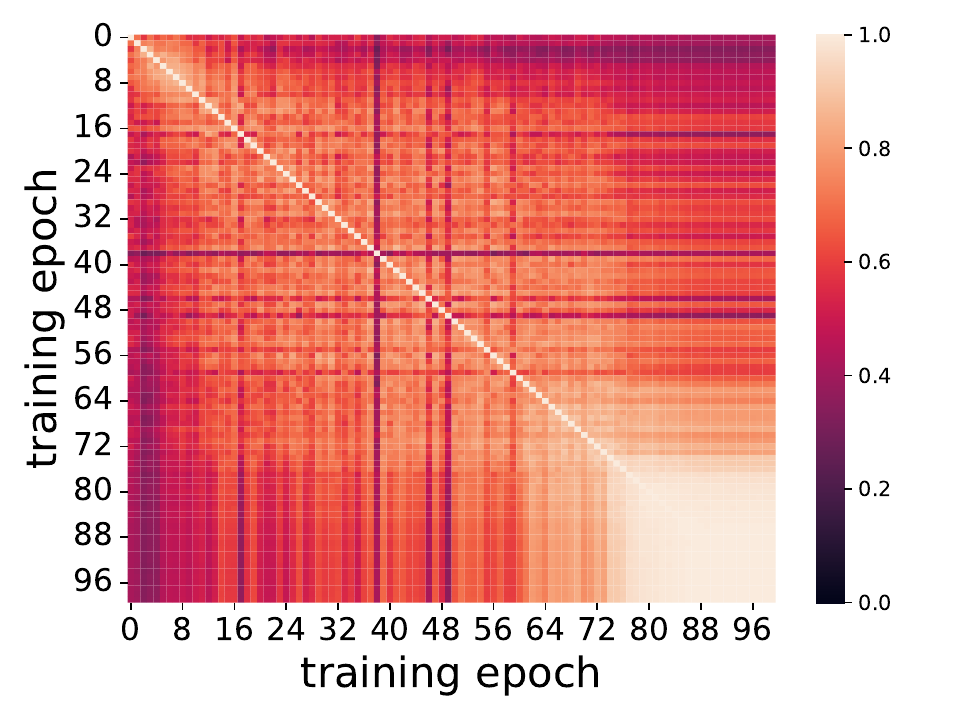}
    \caption{}
    \label{fig:r}
\end{subfigure}
\begin{subfigure}[t]{0.48\columnwidth}
    \centering
    \includegraphics[width=\textwidth]{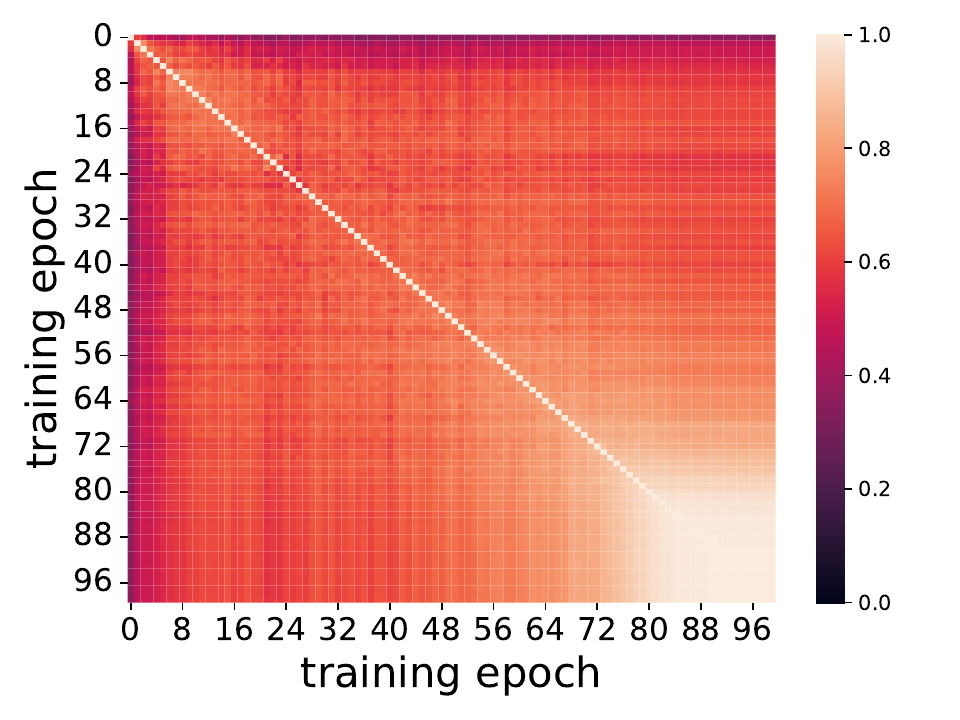}
    \caption{}
    \label{fig:ow}
\end{subfigure}
\begin{subfigure}[t]{0.48\columnwidth}
    \centering
    \includegraphics[width=\textwidth]{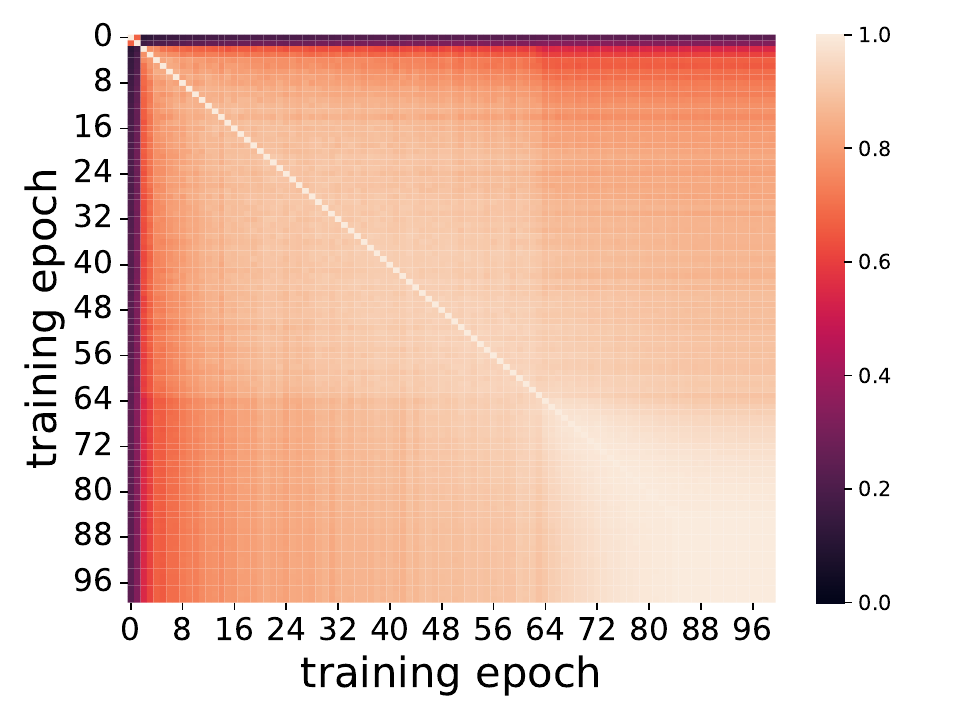}
    \caption{}
    \label{fig:o}
\end{subfigure}
\caption{\textbf{CKA of different models at last convolutional layer over time.} \textbf{(a)} \textit{Target Network:} A smooth pattern that becomes lighter along the diagonal toward the end of training when training from scratch; \textbf{(b)} \textit{Random:} More dark and noisy pattern compared with the target when growing with random weights; \textbf{(c)} \textit{\OurMethodWithoutEns{}:} Without model fusion, the pattern is somewhat similar to the Target, but the representations between epochs have smaller similarities; \textbf{(d)} \textit{\OurMethod{}:} In our full model, we see a clear phase change at the growth point (epoch 63), suggesting rapid learning that occurs after fusing the 2 small models.}
\label{fig:cka_by_epoch}
\end{figure}

\begin{figure}[t]
\centering
\vspace{-2mm}
\begin{subfigure}[t]{0.48\columnwidth}
    \centering
    \includegraphics[width=\textwidth]{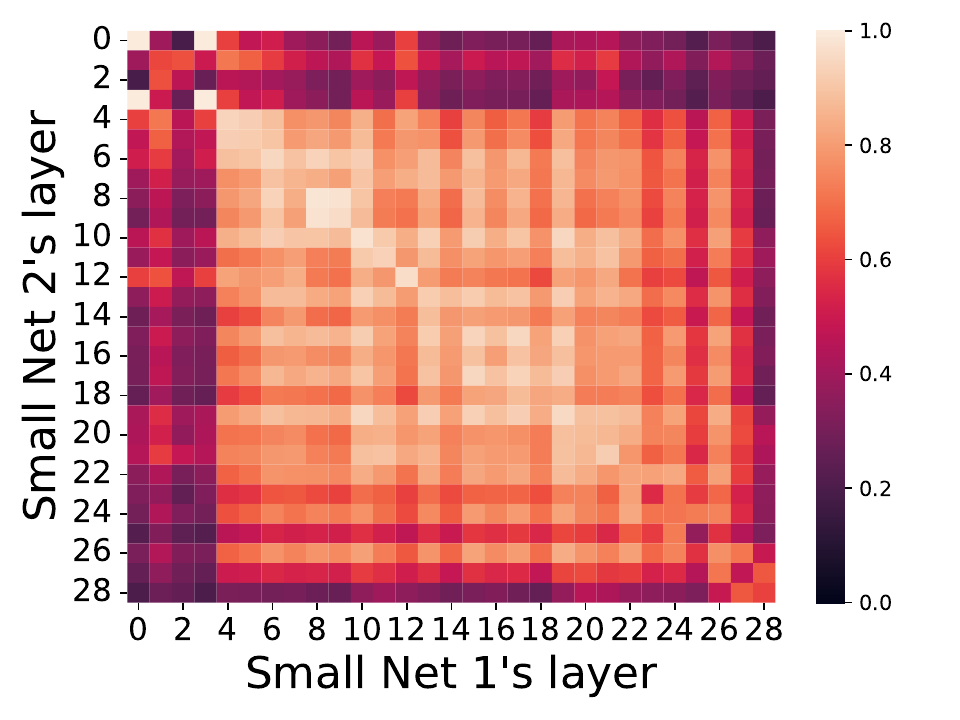}
    \caption{}
    
    \label{fig:sm1_sm2}
    
\end{subfigure}
\begin{subfigure}[t]{0.48\columnwidth}
    \centering
    \includegraphics[width=\textwidth]{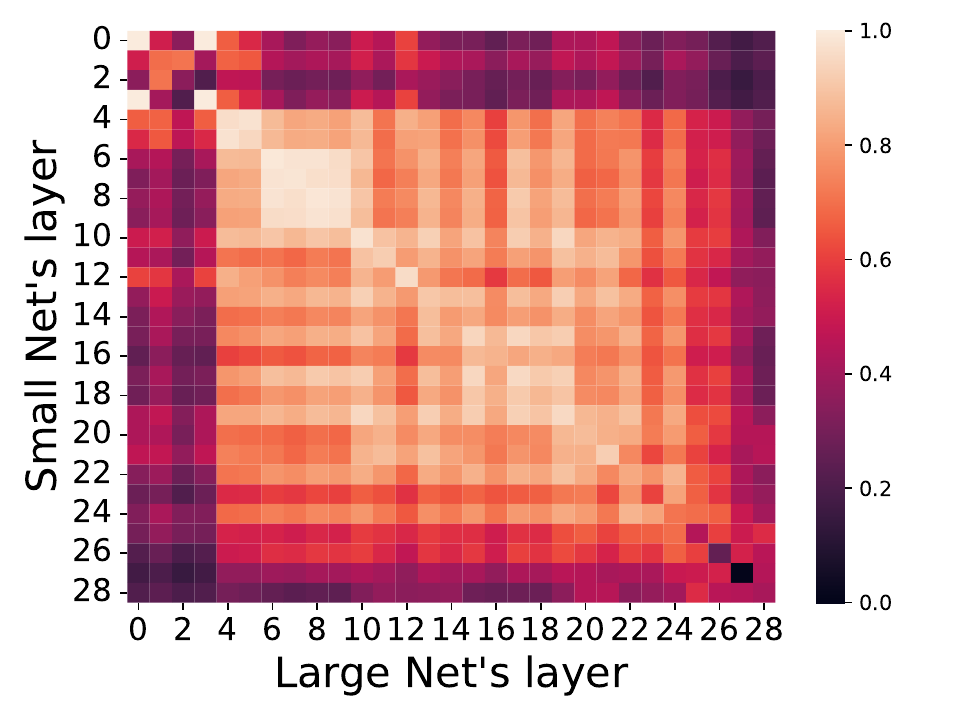}
    \caption{}
    \label{fig:l_sm}
\end{subfigure}
\begin{subfigure}[t]{0.48\columnwidth}
    \centering
    \includegraphics[width=\textwidth]{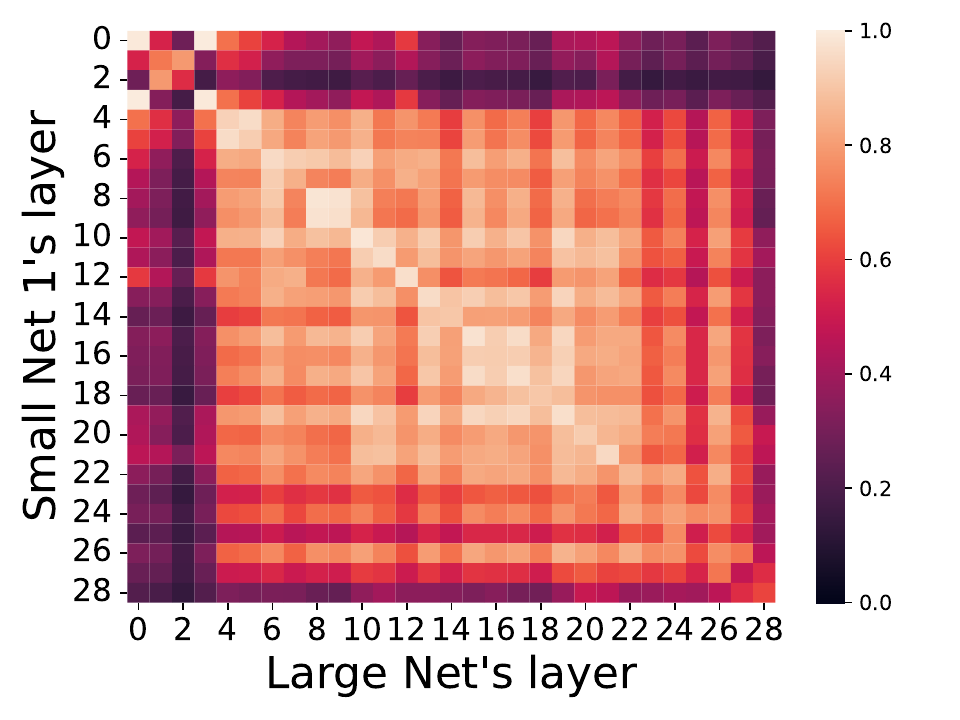}
    \caption{}
    \label{fig:l_sm1}
\end{subfigure}
\begin{subfigure}[t]{0.48\columnwidth}
    \centering
    \includegraphics[width=\textwidth]{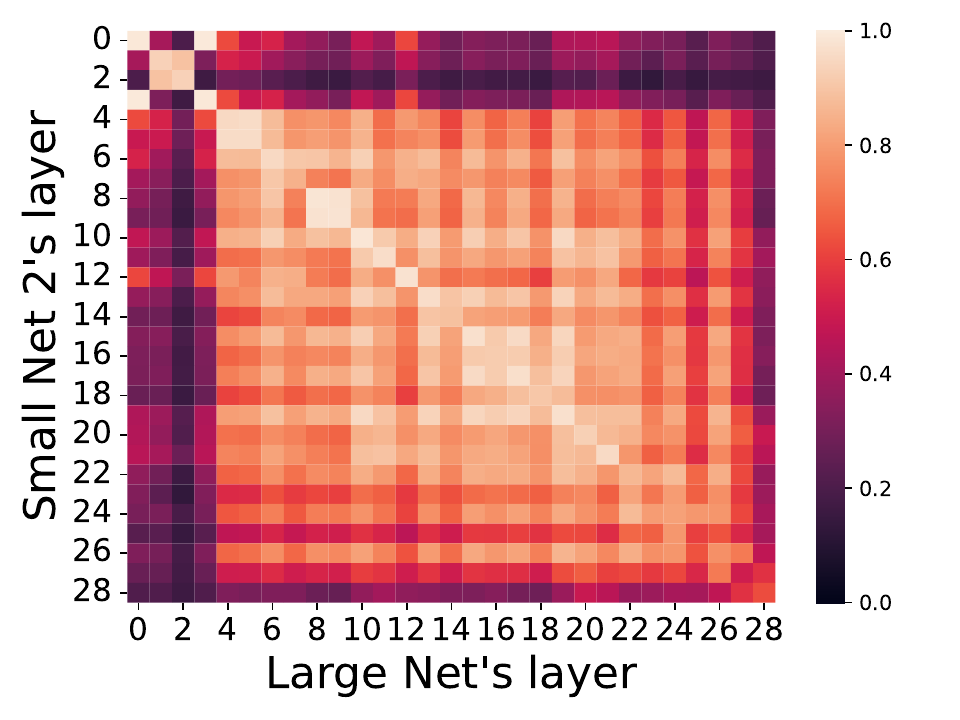}
    \caption{}
    \label{fig:l_sm2}
\end{subfigure}

\caption{\textbf{CKA between layers of \OurMethod{}.} \textbf{(a)} \OurMethod{}:  Small Network 1 vs. Small Network 2; \textbf{(b)} \OurMethod{} w/o model fusion:  Large Network vs. Small Network 1; \textbf{(c)} \OurMethod{} with model fusion:  Large Network vs. Small Network 1; \textbf{(d)} \OurMethod{} with model fusion:  Large Network vs. Small Network 2;  We analyze the similarity of representations for each pair of layers of a WRN-28-10 on CIFAR-100 dataset. 
See Sec.~\ref{subsubsec:feature_analysis} for discussion.}

\label{fig:cka_by_layer}
\end{figure}

We use Centered Kernel Alignment (CKA)~\cite{kornblith2019similarity} to analyze the similarity of features over training. For this analysis, we set up \OurMethod{} by training 2 small networks for 63 epochs, then training the fully grown network afterward until reaching 100 epochs in total. For \OurMethodWithoutEns{}, only one small model was trained for 63 epochs, and then we grew. The target network refers to a full network trained from scratch for 100 epochs. 

Fig.~\ref{fig:cka_by_epoch} shows how features evolve during training over time using CKA procedure described above. In Fig.~\ref{fig:o}, which visualizes feature similarity, a very clear phase change in similarity is visible at the growth step at epoch 63. This indicates that learning at the growth point is very rapid, suggesting we effectively fuse two independent models in the growth step. Random weight initialization (Fig.~\ref{fig:r}) reflects a similar change seen in  Fig.~\ref{fig:o}, but with more noisy patterns. In contrast, training the target network from scratch (Fig.~\ref{fig:t}), and \OurMethodWithoutEns{} (Fig.~\ref{fig:ow}) have a more gradual transition from the beginning of training towards the end without clear quadrants like \OurMethod{}. Thus, learning during growth is not rapid, but gradual instead. This provides insight into the effectiveness of our approach, especially with low FLOP budgets.

Fig.~\ref{fig:cka_by_layer} illustrates how networks have changed from the point of growth to the final representation with CKA. In Fig.~\ref{fig:sm1_sm2}, we compare the two small networks right before growth. We observe that several layers across networks have similar features (light color on the diagonal, which represents highly similar features from the same layer), and some early and late layers differ (dark color).  We find a similar pattern growing from a single model (\ie, \OurMethodWithoutEns{}, Fig.~\ref{fig:l_sm}). This suggests that after growth (w/o fusion), the large network learns additional features that differ from the small net in a similar way as two independently trained small nets. In contrast, in \OurMethod{} w/fusion, from Fig.~\ref{fig:l_sm1}  and \ref{fig:l_sm2}, we see the full net shares more similarity with each of the small nets in the early and late layers than \OurMethodWithoutEns{}. Thus, little learning happens after fusing two small nets, and lightweight training after growth is sufficient.

\section{Conclusion}
\label{sec:conclusion}
In this paper, we propose \OurMethod, a new method to grow networks with weights constructed from linear combinations of shared templates that employs model fusion to boost diversity in our pretrained templates. We analyze our approach by answering several questions about the impact of linear coefficients initialization and finding a good growth point. We find that without model fusion, \OurMethod{} already achieves comparable or higher accuracy than the state-of-the-art. However, with fusion the benefits over prior work increases, resulting in a 2-2.5\% improvement over the state-of-the-art on CIFAR-100 and ImageNet. We believe \OurMethod{} to represent an interesting step forward in learning to grow networks by leveraging template mixing to generate new weights.
\smallskip

\noindent\textbf{Acknowledgements}
This material is based upon work supported, in part, by DARPA under agreement number HR00112020054 and the NSF under award DBI-2134696. Any opinions, findings, and conclusions or recommendations are those of the author(s) and do not necessarily reflect the views of the supporting agencies.

{\small
\bibliographystyle{ieee_fullname}
\bibliography{main}
}

\newpage
\centerline{\huge{\textbf{Appendix}}}
\vskip 1cm
\section{Implementation Details}
We set up our experiments on image classification tasks where the goal is to recognize the object in an image. This is evaluated using top 1 accuracy, \ie, the percentage of times the model can correctly predict the category of an image. We evaluate our method and the baselines on CIFAR-10 and  CIFAR100\cite{Krizhevsky09learningmultiple}, which consists of 60K images of 10 and 100 categories respectively, and ImageNet~\cite{deng2009imagenet}, which contains 1.2M images of 1,000 categories.

\begin{figure*}[t]
\centering
\begin{subfigure}[t]{0.95\columnwidth}
\centering
\includegraphics[width=\linewidth]{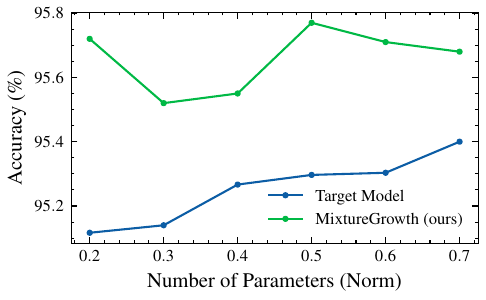}
    \caption{CIFAR-10}
 \label{fig:cifar10_small_model}
\end{subfigure}
\begin{subfigure}[t]{0.95\columnwidth}
 \centering
   \includegraphics[width=\linewidth]{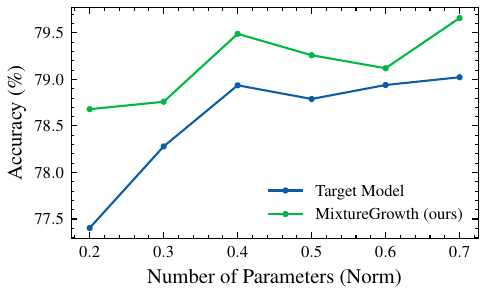}
    \caption{CIFAR-100}
 \label{fig:cifar100_small_model}
\end{subfigure}
\caption{Comparison of \OurMethod{} with target model under low parameter settings. Performance measured by top-1 accuracy averaged over 3 runs.}
\label{fig:low_param}
\end{figure*}

\subsection{CIFAR-10 and CIFAR-100} 
CIFAR-10 and CIFAR100\cite{Krizhevsky09learningmultiple} are composed of 60K images of 10 and 100 categories respectively. In both datasets,
we split the data into 50K images for training and 10K images for testing. We perform experiments using Wide Residual Network (WRN) architecture~\cite{zagoruyko2016wide}, which is a modified version of residual network~\cite{he2016deep}. We denote a WRN model as  WRN-$n$-$k$, where $n$ is the total number of convolutional layers, and $k$ is the widening factor. WRN increases the width of each layer by a factor of $k$ while decreasing the depth to improve the performance of the traditional residual network. In this experiment, we choose WRN-28-10 adapted from Savarese \etal \footnote{\url{https://github.com/lolemacs/soft-sharing}}. The network is trained for 200 epochs in total on a single GPU (Nvidia Titan V 12G) using stochastic gradient descent with momentum 0.9, learning rate 0.1, which is decayed to 0 with a cosine schedule, a weight decay of $5\times10^{-4}$, and a batch size of 128. For the loss function, we use cross entropy loss. To have a fair comparison, we follow the same setting in prior work~\cite{evci2022gradmax}, where Batch Normalization~\cite{IoffeBatchNorm2015} is only used before each block. Parameters that are specific to our method are set as follows. The total parameter budget for template mixing is set to 36.5M, which is equal to the number of parameters in WRN-28-10 (target network). Note that the parameter budget in our method can be flexible, \ie, it can have a setup where the model has fewer or more parameters thanks to template mixing schemes, see Section \ref{sec:low_param_budget} for a comparison of varying of parameter budgets.  For each layer, the number of templates is set to 2. Except for the first (conv 1) and last (Fully connected) layers, all alpha coefficients are trainable. To train \OurMethod{}, given a trained WRN-28-5 model, we begin with training another WRN-28-5 model for $e$ epochs, where $0 \leq e <200$ is a hyper-parameter. Then, these small models are fused and used to initialize for growing to the full model (\ie, WRN-28-10), which is 4x larger in terms of the number of weights. We train the full model for some more epochs, depending on the FLOPs budget. 

It is worth noting that GradMax~\cite{evci2022gradmax} shrinks the first convolutional layer at every block by 4, resulting in having a small network with wide and narrow layers alternately. In contrast, \OurMethod{} reduces both the widths of its input and output in each layer by 2 to achieve a smaller version of the network. Though the small networks in our method and the baselines are slightly different due to the setting of each method, all of them have the same FLOPs (around 0.25X FLOPs Norm) for a fair comparison.

\subsection{ImageNet}
ImageNet~\cite{deng2009imagenet} contains 1,000 categories with 1.2M images for training, 50K for validation, and 100K for testing. We train models using the ResNet-50 architecture\cite{he2016deep} for 90 epochs on 4 GPUs (NVIDIA RTX A6000 48G) with a learning rate of 0.1, which is decayed by 0.1 at 30, 60, and 80 epochs with a cosine scheduler.  We use stochastic gradient descent with a momentum of 0.9, a batch size of 256, and cross entropy as the loss function. In \OurMethod{}, we share templates between two consecutive layers if they have the same size. The total parameter for template mixing in our method is 25.6M, which is equal to that of a target model. It is worth noting that in GradMax~\cite{evci2022gradmax}, the small network they used required more FLOPs than the one utilized in our method (0.3X FLOPs Norm for GradMax versus 0.26X FLOPs Norm for ours), due to the setup of the authors described in the previous section. However, all other settings remain the same as those mentioned for CIFAR above.

To train our method, we start with a trained network whose input and output of each of its layers are half of the sizes of the target network. We train another equal-sized network for $e$ epochs, where  $0 \leq e < 90$ is a hyperparameter. Then, these small networks are fused to grow into a full network. We train the fully grown network for some more epochs until run out of the FLOPs budget.

\begin{figure}[t]
    \centering
    \includegraphics[width=\linewidth]{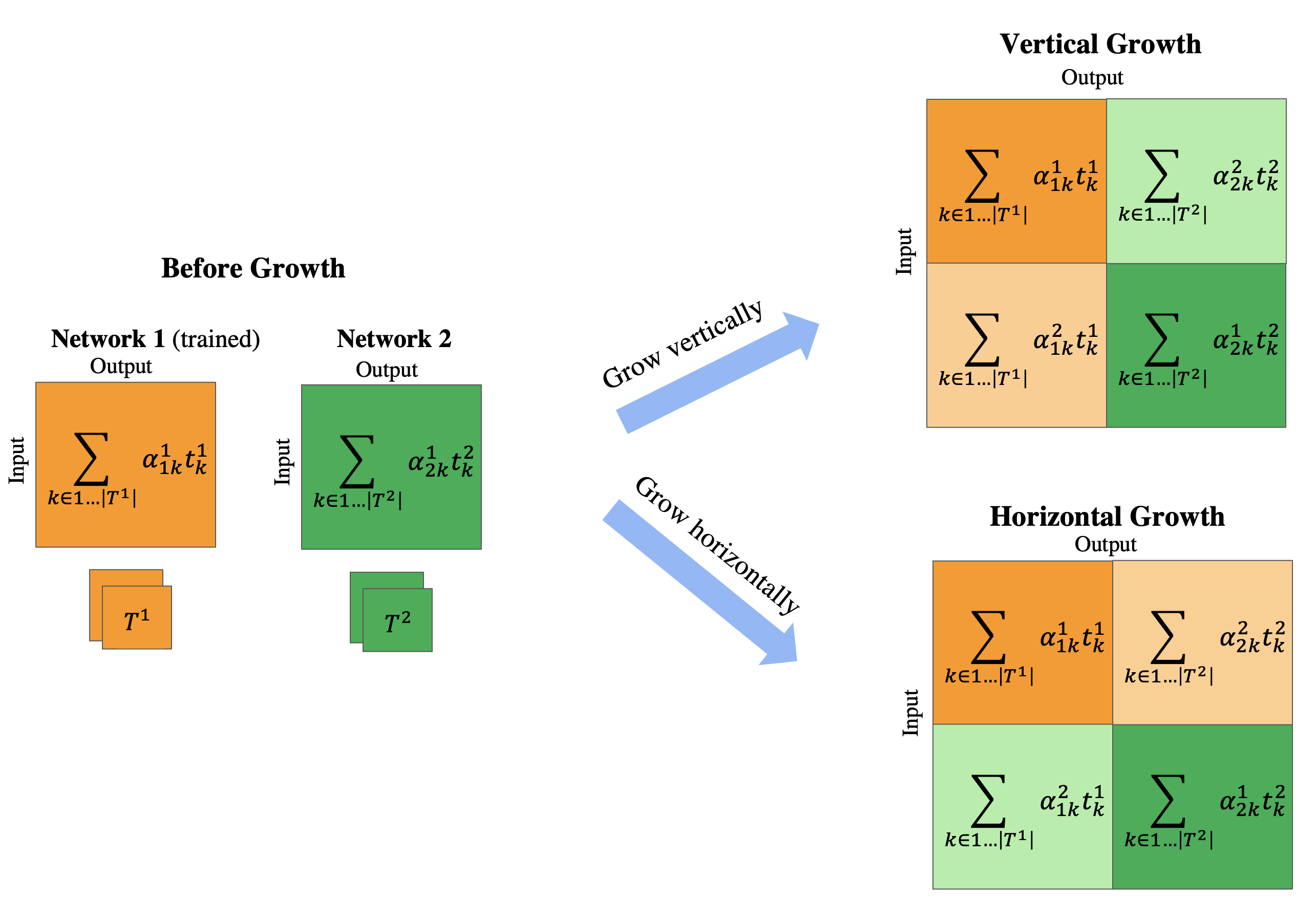}
    \caption{\textbf{\OurMethod{} with different growth methods}: Illustration of Vertical growth and  Horizontal growth. Quadrants that share templates but have different linear combination coefficients are presented as different shades of the same color.}
 \label{fig:vertical_horizontal}
\end{figure}

\section{Experiments with VGG-11}

Besides the WideResnet ~\cite{zagoruyko2016wide}, we conduct experiments on different families of architecture. Table \ref{table:vgg11} shows the performance of our method and the baselines on CIFAR-100 dataset when growing from 2 small networks. Firefly struggles to converge, resulting in being excluded from the table. However, Random and GradMax give similar results, with about 48\% accuracy. Our method performs better than the baselines by a large margin ($\sim$ 7.5\%). 

\begin{table}[t]
\centering
\caption{\textbf{Network growing comparison on CIFAR-100 using VGG-11}\cite{simonyan2014very} architecture. (a) Performance of baseline models where Firefly is not included due to non-convergence (b) Performance of \OurMethod 
}
\resizebox{0.49\textwidth}{!}{

\begin{tabular}{rlcc} 
\toprule
 & Method & Top-1 Acc. &  Total FLOPs Norm   \\
 \midrule
 \textbf{(a)} & Random & 48.52\% &   1.3X \\
 & GradMax~\cite{evci2022gradmax} & 48.79\% &  1.3X \\  

 \midrule
 \textbf{(b)} & \textbf{\OurMethod\ (ours)} & \textbf{56.17\%}   & 0.6X \\
 \bottomrule
\end{tabular}
}

\vspace{6pt}

\label{table:vgg11}
\end{table}


\setlength{\tabcolsep}{8pt} %
\begin{table}[t]
\centering
\caption{Comparison different growth method of \OurMethod{} on CIFAR-100 using WRN-28-10 architecture. We report average accuracy of three runs for each method. 
}
\begin{tabular}{lcc} 
\toprule
 Method & Top-1 Acc. &  FLOPs Norm   \\
 \midrule
 Horizontal growth & 80.66\% &  0.35X \\
 Vertical growth & \textbf{80.82}\% &  0.35X \\

 \bottomrule
\end{tabular}

\label{table:horizontal_vertical}
\end{table}



\section{Low Parameter budgets}
\label{sec:low_param_budget}
Template mixing allows us to share parameters across layers, reducing the number of parameters in the network without the need to change its architecture (such as width and depth). To compare our method with target models under low parameter budgets, we reduce the width of target models so that their number of parameters matches that of our method. Figure \ref{fig:cifar10_small_model} illustrates the performance of \OurMethod{} when compared with small target models on CIFAR-10 with the same number of parameters, where \OurMethod{} consistently outperforms the target models under low parameter budgets. We find a similar observation on CIFAR-100 dataset, as shown in \ref{fig:cifar100_small_model}. We use WRN architecture~\cite{zagoruyko2016wide} for the comparison of both datasets.


\section{Horizontal and Vertical Growth}

At the growth step, we grow from 2 trained small networks into a large network. Given the trained networks are the 2 diagonal quadrants, there are 2 ways to expand it into 4 quadrants. The first option is Vertical growth, where quadrants that have the same output share the templates (Figure \ref{fig:vertical_horizontal}, top right). The other way is horizontal growth in which quadrants that have the same input use the same set of templates (Figure \ref{fig:vertical_horizontal}, bottom right). Table \ref{table:horizontal_vertical} compares the performance of the 2 growth strategies on CIFAR-100 dataset. We notice that Vertical growth slightly outperforms Horizontal growth in terms of performance.

\end{document}